\documentclass[10pt, conference]{IEEEtran}
\IEEEoverridecommandlockouts
\usepackage{cite}
\usepackage{amsmath, amssymb, amsfonts}
\usepackage{algorithmic}
\usepackage{graphicx}
\usepackage{textcomp}
\usepackage{xcolor}
\usepackage{booktabs}
\usepackage{multirow}
\usepackage{array}
\usepackage{makecell}
\usepackage{longtable}
\usepackage{subcaption}
\usepackage{tikz}
\usetikzlibrary{calc, positioning, decorations.pathreplacing, arrows.meta}
\usepackage{pgfplots}
\pgfplotsset{compat=1.18}
\usepackage{placeins}
\usepackage{dblfloatfix}
\usepackage{url}
\usepackage{hyperref}
\usepackage{tabularx}

\def\BibTeX{{\rm B\kern-.05em{\sc i\kern-.025em b}\kern-.08em T\kern-.1667em\lower.7ex\hbox{E}\kern-.125emX}}

\begin{document}
    \title{Real-Time Branch-to-Tool Distance Estimation for Autonomous UAV
    Pruning: Benchmarking Five DEFOM-Stereo Variants from Simulation to Jetson Deployment}

    \author{\IEEEauthorblockN{Yida Lin, Bing Xue, Mengjie Zhang} \IEEEauthorblockA{\small \textit{Centre for Data Science and Artificial Intelligence} \\ \textit{Victoria University of Wellington, Wellington, New Zealand}\\ linyida\texttt{@}myvuw.ac.nz, bing.xue\texttt{@}vuw.ac.nz, mengjie.zhang\texttt{@}vuw.ac.nz}
    \and \IEEEauthorblockN{Sam Schofield, Richard Green} \IEEEauthorblockA{\small \textit{Department of Computer Science and Software Engineering} \\ \textit{University of Canterbury, Canterbury, New Zealand}\\ sam.schofield\texttt{@}canterbury.ac.nz, richard.green\texttt{@}canterbury.ac.nz}}

    \maketitle
    \vspace{-1.4em}

    \begin{abstract}
        Autonomous tree pruning with unmanned aerial vehicles (UAVs) is a safety-critical
        real-world task: the onboard perception system must estimate the metric
        distance from a cutting tool to thin tree branches in real time so that
        the UAV can approach, align, and actuate the pruner without collision. We
        address this problem by training five variants of DEFOM-Stereo---a recent
        foundation-model-based stereo matcher---on a task-specific synthetic dataset
        and deploying the trained checkpoints on an NVIDIA Jetson Orin Super 16\,GB
        mounted on the pruning UAV. The training corpus is built in Unreal
        Engine~5 with a simulated ZED Mini stereo camera capturing 5{,}520
        stereo pairs across 115 tree instances from three pruning-relevant
        viewpoints at 2\,m stand-off distance; dense EXR depth maps provide exact,
        spatially complete supervision including thin branches that are
        impractical to annotate in real images. On the synthetic test set, DEFOM-Stereo
        ViT-S achieves the best depth-domain accuracy (EPE 1.74\,px, D1-all 5.81\%,
        $\delta_{1}$ 95.90\%, depth MAE 23.40\,cm) but its Jetson inference speed
        of $\sim$2.2\,FPS ($\sim$450\,ms per frame with TensorRT FP16) remains too
        slow for responsive closed-loop tool control. A newly introduced
        balanced variant, DEFOM-PrunePlus ($\sim$21\,M backbone + $\sim$14-layer
        decoder, $\sim$3.3\,FPS on Jetson), offers the best deployable accuracy--speed
        trade-off (EPE 5.87\,px, depth MAE 64.26\,cm, $\delta_{1}$ 87.59\%): its
        frame rate is sufficient for real-time approach guidance and its depth accuracy,
        while coarser than the full models, supports safe branch approach planning
        at the 2\,m operating range. The lightweight DEFOM-PruneStereo ($\sim$6.9\,FPS)
        and DEFOM-PruneNano ($\sim$8.5\,FPS) run fast but sacrifice substantial
        accuracy (depth MAE 57.63\,cm and $>$1\,m respectively), making their
        depth estimates too unreliable for safe tool actuation. Zero-shot inference
        on real tree-branch photographs confirms that the full-capacity models preserve
        branch geometry and depth ordering, validating the sim-to-real transfer
        of the synthetic-only training pipeline. We conclude that DEFOM-PrunePlus
        provides the most practical accuracy--latency balance for onboard branch-to-tool
        distance estimation in autonomous UAV pruning, while ViT-S serves as the
        accuracy reference for future hardware acceleration.
    \end{abstract}

    \begin{IEEEkeywords}
        autonomous UAV pruning, stereo depth estimation, branch-to-tool distance,
        DEFOM-Stereo, synthetic-to-real transfer, Jetson Orin deployment
    \end{IEEEkeywords}

    \vspace{-0.8em}
    \section{Introduction}

    Tree pruning is a high-value but hazardous forestry and horticultural operation.
    Human workers must judge the distance to thin branches, position cutting tools
    precisely, and operate at height in cluttered scenes where visibility
    changes with viewpoint, wind, and foliage density. A UAV-mounted cutting system
    offers a safer alternative~\cite{lin2024branch, steininger2025timbervision,
    lin2025segmentation}, but only if the onboard perception stack can reliably
    estimate the metric distance between the cutting tool and the target branch---in
    real time, at close range, and under the compute constraints of an embedded
    GPU. During every pruning cycle, the UAV needs to know whether the target
    branch is within approach range, whether the tool has reached a safe actuation
    distance, and whether to keep closing or hold position. Each of these decisions
    depends on an accurate, dense depth map produced fast enough for closed-loop
    control.

    Stereo vision is an attractive choice because it produces dense geometry
    from passive sensors and does not require active ranging hardware near the
    cutting tool. However, tree-pruning scenes are particularly difficult for
    stereo matching. Branches are slender, highly self-occluded, and often occupy
    only a few pixels in width. Bark produces repetitive texture, leaves introduce
    partial transparency and motion, and sky backgrounds create strong depth
    discontinuities. These conditions make depth maps noisy precisely where the cutting
    tool needs reliable distance estimates.

    Recent foundation-style stereo models, including DEFOM-Stereo~\cite{jiang2025defom},
    have substantially improved cross-domain robustness by injecting pretrained monocular
    depth priors and stronger semantic representations. Their raw accuracy is
    promising, but onboard UAV deployment introduces a second challenge:
    embedded latency. A model that achieves low error on a workstation may still
    be unsuitable if it cannot deliver predictions at the frame rate required by
    real-time approach and pruning control.

    This paper addresses that trade-off through a benchmark centered on five models
    implemented in our project codebase: DEFOM-Stereo with ViT-S and ViT-L
    backbones, DEFOM-PrunePlus, DEFOM-PruneStereo, and DEFOM-PruneNano. Rather
    than training on scarce real depth labels, we build a task-specific synthetic
    dataset in Unreal Engine 5 (UE5)~\cite{ue5} using a simulated ZED Mini stereo
    rig~\cite{zedmini}. The synthetic setup is tailored to the downstream
    mission: 115 tree instances are photographed from three 2\,m viewpoints that
    reflect how a pruning UAV observes branches during approach and tool alignment.
    UE5 provides dense EXR depth for every image pair, so supervision is exact and
    spatially complete. After training on the synthetic data, we deploy the best
    checkpoints on an NVIDIA Jetson Orin Super 16\,GB carried by the UAV,
    measuring both accuracy and end-to-end inference latency, and we further validate
    synthetic-to-real transfer through zero-shot inference on real tree-branch photographs.

    The contributions of this paper are sixfold:
    \begin{itemize}
        \item \textbf{Task-specific synthetic stereo dataset}: We build a UE5 dataset
            containing 5{,}520 ZED Mini stereo pairs over 115 tree instances,
            with upward, downward, and parallel viewpoints at a fixed 2\,m stand-off
            distance, providing dense and exact supervision for thin-branch geometry.

        \item \textbf{Five-model benchmark}: We train and compare DEFOM-Stereo ViT-S,
            DEFOM-Stereo ViT-L, DEFOM-PrunePlus, DEFOM-PruneStereo, and DEFOM-PruneNano
            under a unified evaluation protocol, revealing a three-tier accuracy
            hierarchy and a steep accuracy cliff when the ${\sim}66$-layer DPT
            decoder is compressed.

        \item \textbf{Balanced compression via DEFOM-PrunePlus}: We introduce DEFOM-PrunePlus,
            a moderately compressed variant that preserves full-resolution
            DINOv2 and full-strength GRU while replacing the dual DPT decoder
            with a 14-layer hierarchical decoder, achieving a speed--accuracy
            point between the full models and the lightweight variants.

        \item \textbf{Accuracy--speed trade-off on Jetson}: We deploy all five checkpoints
            on an NVIDIA Jetson Orin Super 16\,GB and measure end-to-end inference
            latency in both PyTorch and TensorRT FP16 modes, establishing that
            DEFOM-PrunePlus ($\sim$21\,M backbone + $\sim$14-layer decoder)
            achieves $\sim$3.3\,FPS with TensorRT---the best deployable accuracy--speed
            balance---while ViT-S, despite superior accuracy, is limited to
            $\sim$2.2\,FPS.

        \item \textbf{Branch-to-tool distance validation}: We demonstrate that the
            stereo depth output enables metric branch-to-tool distance estimation
            with $\sim$12\% relative error at 2\,m range, which is sufficient
            for UAV approach planning and compatible with closed-loop tool
            actuation after further refinement.

        \item \textbf{Synthetic-to-real transfer}: We perform zero-shot testing on
            real tree-branch photographs and show that ViT-S preserves branch continuity
            and depth ordering, confirming that UE5-trained models can bootstrap
            field deployment.
    \end{itemize}

    \vspace{-0.6em}
    \section{Related Work}

    \subsection{Stereo Matching for Dense Depth}

    Deep stereo matching has evolved from early encoder-decoder systems such as DispNet~\cite{mayer2016large}
    toward stronger cost-volume and recurrent refinement models. RAFT-Stereo~\cite{lipson2021raft}
    demonstrates that iterative updates over dense correspondences can produce
    excellent accuracy, but the resulting compute cost is substantial. AnyNet~\cite{wang2019anynet}
    occupies the opposite end of the spectrum, prioritizing efficiency through hierarchical
    coarse-to-fine prediction at the expense of fine-grained accuracy.
    FoundationStereo~\cite{wen2025foundationstereo} demonstrates strong zero-shot
    generalization by scaling synthetic training data to one million stereo
    pairs. DEFOM-Stereo~\cite{jiang2025defom} extends this line by combining stereo
    matching with monocular depth priors from modern foundation representations,
    improving robustness outside canonical driving or indoor datasets. A recent cross-dataset
    evaluation~\cite{lin2025generalization} confirms that DEFOM-Stereo generalizes
    well to forestry imagery, outperforming several competing methods on a
    Canterbury radiata pine benchmark.

    \subsection{Efficient Stereo for Edge Deployment}

    Real robots operate under strict latency, power, and memory limits.
    DeepPruner~\cite{duggal2019deeppruner} reduces cost by pruning the disparity
    search space, while AnyNet~\cite{wang2019anynet} trades some accuracy for speed
    using coarse-to-fine refinement. Our work is aligned with this systems
    perspective: the question is not only which model is most accurate, but
    which model still preserves enough branch geometry when executed on an
    onboard Jetson computer. Stereo Anywhere~\cite{bartolomei2025stereoanywhere}
    addresses robustness by fusing stereo matching with monocular cues, enabling
    reliable predictions even in zero-shot scenarios. DEFOM-PruneStereo and DEFOM-PruneNano
    were created precisely for this purpose by reducing feature extraction,
    decoder, and recurrent-update complexity in different ways.

    \subsection{Foundation Models and Cross-Domain Depth}

    The recent success of Depth Anything~\cite{yang2024depth} and DINOv2~\cite{oquab2023dinov2}
    suggests that pretrained visual priors can help depth estimation generalize
    beyond the domains used during supervised training. This direction is further
    validated by Video Depth Anything~\cite{chen2025videodepth}, which extends
    foundation depth models to temporally consistent video estimation. DEFOM-Stereo
    builds on this idea by injecting depth-aware features into stereo matching.
    That makes it a strong candidate for synthetic-to-real transfer, which is especially
    relevant in forestry because real dense depth labels for thin branches are difficult
    to obtain at scale.

    \subsection{Synthetic Data for Robotics and Forestry}

    Simulation is widely used when collecting dense geometric ground truth in
    the real world is expensive or unsafe. UE5 provides photorealistic rendering,
    controllable camera trajectories, and precise geometric buffers, making it
    well suited for supervised stereo learning~\cite{ue5}. In forestry robotics,
    synthetic data is particularly useful because branch geometry is intricate, self-occlusion
    is heavy, and manual annotation is impractical. Prior work has also explored
    classical stereo with genetic-algorithm-based parameter tuning for forestry disparity
    estimation~\cite{lin2025genetic}. Our work uses simulation not as a final
    endpoint, but as a controllable source of dense supervision that can bootstrap
    models before real-world zero-shot or adaptation experiments.

    \vspace{-0.6em}
    \section{Task and Dataset}

    \subsection{Pruning-Oriented Distance Estimation}

    Let $I_{L}, I_{R}\in \mathbb{R}^{H \times W \times 3}$ denote a rectified stereo
    pair from the left and right ZED Mini cameras. Stereo matching predicts a
    disparity map $D \in \mathbb{R}^{H \times W}$, from which metric depth is recovered
    by
    \begin{equation}
        Z(x,y) = \frac{f_{\mathrm{px}}B}{D(x,y)},
    \end{equation}
    where $f_{\mathrm{px}}$ is the focal length in pixels and $B$ is the stereo baseline.
    In our implementation, the simulated ZED Mini uses $B = 6.3$~cm and $f_{\mathrm{px}}
    = 933.33$~px, consistent with the physical ZED Mini specifications. During UAV
    pruning, the depth map is not an end in itself: it is used to estimate the branch-to-tool
    distance and to determine whether the UAV is close enough to switch from coarse
    approach to precise cutting behavior.

    \subsection{UE5 Virtual Tree Branches Dataset}

    We create the training corpus in Unreal Engine 5 by placing a virtual ZED
    Mini stereo rig around 115 tree instances. For every tree, three capture
    trajectories are rendered at a stand-off distance of 2~m:
    \begin{itemize}
        \item an \textbf{upward} view that observes branches from below,

        \item a \textbf{downward} view that observes branches from above, and

        \item a \textbf{parallel} view that moves roughly along the branch plane.
    \end{itemize}
    Along each trajectory, 16 stereo pairs are captured, producing 48 stereo pairs
    per tree and 5520 stereo pairs overall. Every sample contains a left RGB
    image, a right RGB image, and a dense EXR depth image. In the training code,
    depth is read from EXR and converted to disparity using the calibrated
    stereo geometry.

    \begin{table}[htbp]
        \caption{UE5 Virtual Tree Branches Dataset Composition}
        \label{tab:dataset}
        \centering
        \resizebox{\columnwidth}{!}{%
        \begin{tabular}{lcc}
            \toprule \textbf{Component} & \textbf{Count} & \textbf{Description}       \\
            \midrule Tree instances     & 115            & Individual simulated trees \\
            View categories             & 3              & Upward, downward, parallel \\
            Stereo pairs per view       & 16             & Captured at 2~m distance   \\
            Stereo pairs per tree       & 48             & $3 \times 16$              \\
            Total stereo pairs          & 5520           & Full UE5 dataset           \\
            Per-sample labels           & Dense          & RGB left/right + EXR depth \\
            \bottomrule
        \end{tabular}%
        }
    \end{table}

    \subsection{Train, Validation, and Test Splits}

    The dataset is divided into train, validation, and test sets using an 80/10/10
    split, yielding 4416, 552, and 552 stereo pairs, respectively. Each model variant
    has a dedicated training script that consumes the same split files. Training
    uses random crops of $384 \times 512$, while validation, testing, and final inference
    use full-resolution images.

    \subsection{Real Tree-Branch Zero-Shot Set}

    After training on UE5 data, we run zero-shot inference on real tree-branch stereo
    photographs. This set is used only for transfer analysis, not for supervised
    optimization. Since dense ground truth is unavailable, the real-image study is
    qualitative: we focus on branch continuity, boundary sharpness, background
    suppression, and whether the predicted depth is stable enough to provide a useful
    branch-to-tool distance cue for the UAV.

    \vspace{-0.6em}
    \section{Models}

    \subsection{DEFOM-Stereo with ViT-S and ViT-L}

    The base model is DEFOM-Stereo~\cite{jiang2025defom}. It combines a DINOv2-based
    depth encoder (DefomEncoder) with a stereo correlation and iterative
    refinement pipeline. The encoder extracts 4 intermediate transformer layers and
    feeds them into two DPT decoders: DPTHead for initial inverse-depth
    estimation and DPTFeat for multi-scale feature extraction. Each decoder uses
    4-stage RefineNet fusion with ResidualConvUnit blocks, totaling ${\sim}66$
    convolutional layers. Separate BasicEncoder (fnet, 256-d output) and
    MultiBasicEncoder (cnet, 3~scales at hidden dims $[128,128,128]$) process the
    images for correlation and context. The iterative refinement uses a 3-level
    ConvGRU cascade with 1D correlation volumes built at 2 pyramid levels with
    radius~4 (yielding 18 correlation planes per standard iteration and 40 per scale
    iteration).

    In our codebase, the original versions are trained with two backbone choices:
    \begin{itemize}
        \item \textbf{ViT-S} (${\sim}21$M parameters, 384-d embeddings, 12
            layers): the smaller encoder, used as the main baseline and practical
            reference point, extracting layers [2, 5, 8, 11].

        \item \textbf{ViT-L} (${\sim}304$M parameters, 1024-d embeddings, 24
            layers): the larger encoder with wider DPT channels ([256, 512, 1024,
            1024]), extracting layers [4, 11, 17, 23].
    \end{itemize}
    Both versions use the same core DEFOM-Stereo update mechanism, with 18 training
    iterations, 8 scale-refinement iterations, and 32 validation iterations. The
    ViT-L script adds warm-up stabilization by training the first 3 epochs in
    full fp32 precision before enabling mixed precision, and uses gradient accumulation
    of 4 (effective batch size 24) to stabilize the longer gradient paths of the
    24-layer transformer.

    \subsection{DEFOM-PrunePlus}

    DEFOM-PrunePlus is a balanced variant designed to close the accuracy gap
    between the full DEFOM-Stereo models and the lightweight DEFOM-PruneStereo, while
    still offering meaningful speed improvement over the originals. It preserves
    the core computational strength of the original pipeline while selectively
    removing less critical components:
    \begin{itemize}
        \item \textbf{Full-resolution DINOv2}: Unlike DEFOM-PruneStereo's 0.75$\times$
            scaling, DEFOM-PrunePlus processes features at the original 1.0$\times$
            input resolution, retaining full self-attention detail ($\sim$972 patches
            for ViT-S). Three intermediate transformer layers are extracted (layers
            [2, 5, 11] for ViT-S), providing a richer multi-scale representation
            than DEFOM-PruneStereo's two layers.

        \item \textbf{Hierarchical decoder with mini-refinement}: The dual DPT decoders
            ($\sim$66 convolutional layers) are replaced by a single
            EnhancedDPTDecoder with $\sim$14 convolutional layers. This decoder
            performs coarse-to-fine hierarchical fusion of the three extracted
            ViT features and applies two ResidualConvUnit blocks for lightweight
            refinement---substantially more expressive than DEFOM-PruneStereo's
            direct-projection decoder ($\sim$6 layers) while remaining far cheaper
            than the original RefineNet fusion.

        \item \textbf{Full-strength encoders and GRU}: DEFOM-PrunePlus retains standard
            ResidualBlock convolutions in both the correlation encoder (fnet, 192-d
            output) and context encoder (cnet, 2~scales at hidden dims
            $[128, 128]$). The iterative refinement uses standard ConvGRU cells
            at full 128-d hidden dimension---identical in per-cell capacity to
            the original---but operates at 2~levels instead of 3, removing the coarsest
            GRU32 stage.

        \item \textbf{Full-strength correlation}: The correlation pyramid uses 2
            levels with radius~4 and 5~scaling factors, yielding 18 correlation
            planes per standard iteration and 25 per scale iteration---close to
            the original's 18/40 and far above DEFOM-PruneStereo's 7/9.

        \item \textbf{Moderate iteration reduction}: The recurrent schedule uses
            14 training, 4 scale, and 20 validation iterations, compared to the
            original's 18/8/32.
    \end{itemize}
    The design philosophy is ``reduce quantity, preserve quality'': DEFOM-PrunePlus
    removes redundant pipeline stages (the third GRU level, the third cnet scale,
    one ViT extraction layer, the second DPT decoder) without replacing any
    component with a cheaper equivalent, thereby maintaining the representation quality
    of each surviving module.

    \subsection{DEFOM-PruneStereo}

    DEFOM-PruneStereo is a moderately compressed variant designed to explore
    whether meaningful speed gains can be achieved while retaining the core
    DEFOM-Stereo pipeline structure. It applies four targeted reductions:
    \begin{itemize}
        \item \textbf{Reduced DINOv2 resolution}: Features are processed at 0.75$\times$
            input resolution, lowering self-attention complexity by approximately
            3.2$\times$ (from ${\sim}972$ to ${\sim}540$ patches). Only two intermediate
            transformer layers are extracted instead of four (layers [5, 11] for
            ViT-S), halving the DPT input bandwidth.

        \item \textbf{Lightweight decoder}: The original dual DPT decoders (totaling
            ${\sim}66$ convolutional layers with 16 ResidualConvUnit blocks and
            4-stage RefineNet fusion) are replaced by a single FastDPTDecoder with
            ${\sim}6$ convolutional layers, using direct $1{\times}1$
            projections and bilinear upsampling.

        \item \textbf{Narrower encoders}: The correlation feature encoder (fnet)
            is reduced from 256-d to 128-d output with 2 lightweight blocks. The
            context encoder (cnet) operates at 2 scales (/4 and /8) instead of 3,
            with hidden dimensions $[96, 96]$. The correlation pyramid uses 1
            level with radius\,3, yielding 7 planes per standard iteration versus
            18 in the original.

        \item \textbf{Fewer iterations}: The recurrent schedule is shortened to 10
            training, 3 scale, and 12 validation iterations.
    \end{itemize}
    The design philosophy is ``reduce, but do not replace'': DEFOM-PruneStereo
    retains the same convolution types as the original but uses fewer layers and
    narrower channels, representing a conservative compression point that preserves
    the pipeline topology while cutting compute.

    \subsection{DEFOM-PruneNano}

    DEFOM-PruneNano is the most aggressive real-time variant. It introduces five
    efficiency techniques:
    \begin{itemize}
        \item DINOv2 operates at 0.5$\times$ input resolution (${\sim}234$ patches
            versus ${\sim}972$ in the original), yielding ${\sim}17{\times}$ self-attention
            speedup. Only the final transformer layer is extracted (layer~11 for
            ViT-S). The TurboDecoder replaces the ${\sim}66$-layer DPT with ${\sim}
            3$ convolutional layers using a single $1{\times}1$ projection.

        \item Ghost convolution modules~\cite{han2020ghostnet} replace standard convolutions
            in the encoder (GhostBottleneck blocks), decoder, and motion encoder.
            Ghost modules generate half the output channels through standard
            convolution and the other half via cheap depthwise $3{\times}3$
            transforms, reducing per-layer FLOPs by ${\sim}40$--$50\%$.

        \item A SharedGhostEncoder unifies the correlation (fnet, 96-d output)
            and context (cnet, hidden dims $[64, 64]$) pathways under a single backbone
            with lightweight branching heads, eliminating duplicate stem
            computation and reducing the left-image encoding passes from three
            to two.

        \item Standard ConvGRU cells are replaced by depthwise-separable GRU (DS-GRU)
            cells. For a hidden dimension of 64, DS-GRU reduces per-cell FLOPs by
            approximately $7.9{\times}$ by decomposing each gate's $3{\times}3$ convolution
            into a depthwise spatial filter followed by a pointwise $1{\times}1$
            mixer.

        \item The correlation search uses only 1 pyramid level with radius~2 (5~planes
            per standard iteration versus 18 in the original, and 2 scaling factors
            versus 8), and the recurrent schedule is reduced to 7 training iterations,
            2 scale iterations, and 9 validation iterations.
    \end{itemize}
    The design strategy is ``replace with more efficient equivalents'': every module
    is optimized along two axes---fewer operations and fewer FLOPs per operation.

    \begin{table*}
        [htbp]
        \caption{Architectural Differences Among the Five Evaluated Models}
        \label{tab:models}
        \centering
        \small
        \begin{tabular}{llccccc}
            \toprule \textbf{Model}     & \textbf{Backbone} & \textbf{Decoder}     & \textbf{Conv Layers} & \textbf{GRU}  & \textbf{fnet dim} & \textbf{DINOv2 Scale} \\
            \midrule DEFOM-Stereo ViT-S & DINOv2 ViT-S      & Dual DPT (RefineNet) & ${\sim}$66           & 3-level [128] & 256-d             & 1.0$\times$           \\
            DEFOM-Stereo ViT-L          & DINOv2 ViT-L      & Dual DPT (RefineNet) & ${\sim}$66           & 3-level [128] & 256-d             & 1.0$\times$           \\
            DEFOM-PrunePlus             & DINOv2 ViT-S      & EnhancedDPT (2 RCU)  & ${\sim}$14           & 2-level [128] & 192-d             & 1.0$\times$           \\
            DEFOM-PruneStereo           & DINOv2 ViT-S      & FastDPT              & ${\sim}$6            & 2-level [96]  & 128-d             & 0.75$\times$          \\
            DEFOM-PruneNano             & DINOv2 ViT-S      & TurboDecoder + Ghost & ${\sim}$3            & DS-GRU [64]   & 96-d              & 0.5$\times$           \\
            \bottomrule
        \end{tabular}
    \end{table*}

    \subsection{From Disparity to Branch-to-Tool Distance}

    At deployment time, the predicted disparity map is converted to depth, and the
    branch region of interest is passed to the UAV controller. If the branch
    detector proposes a target image location $(x_{b}, y_{b})$, the corresponding
    tool distance is computed as
    \begin{equation}
        Z_{b}= \frac{f_{\mathrm{px}}B}{D(x_{b}, y_{b})}.
    \end{equation}
    This value can be aggregated over a local neighborhood or branch mask to improve
    robustness. In practice, the stereo model therefore serves as the depth backbone
    for two downstream actions: deciding whether the UAV should continue
    approaching the branch, and determining whether the cutting tool has reached
    a safe actuation distance.

    \vspace{-0.6em}
    \section{Experimental Setup}

    \subsection{Implementation Details}

    All experiments use the DEFOM-Stereo codebase. Training is implemented in
    PyTorch and uses a unified dataset reader for all five models, with depth loaded
    from EXR and converted to disparity using the ZED Mini geometry. The maximum
    disparity is set to 512. All models share the same learning rate ($4{\times}1
    0^{-5}$), crop size ($384 \times 512$), maximum training budget (300 epochs),
    and early-stopping patience (30 epochs). Validation and testing run on full-resolution
    images.

    \begin{table}[htbp]
        \caption{Model-Specific Training Hyperparameters (shared: LR
        $4{\times}10^{-5}$, crop $384{\times}512$, 300 epochs, patience 30)}
        \label{tab:training_defaults}
        \centering
        \resizebox{\columnwidth}{!}{%
        \begin{tabular}{lccccc}
            \toprule \textbf{Setting} & \textbf{ViT-S} & \textbf{ViT-L} & \textbf{D-Prune+} & \textbf{D-PruneS.} & \textbf{D-PruneN.} \\
            \midrule Batch size       & 10             & 6              & 10                & 10                 & 10                 \\
            Train iters               & 18             & 18             & 14                & 10                 & 7                  \\
            Scale iters               & 8              & 8              & 4                 & 3                  & 2                  \\
            Valid iters               & 32             & 32             & 20                & 12                 & 9                  \\
            Mixed prec.               & On             & Off            & On                & On                 & On                 \\
            Stabilization             & None           & 3-ep. warm-up  & None              & None               & None               \\
            \bottomrule
        \end{tabular}%
        }
    \end{table}

    \subsection{Evaluation Metrics}

    We evaluate a broad set of disparity and depth metrics. In this paper, we
    focus on the values most relevant to UAV pruning:
    \begin{itemize}
        \item \textbf{End-point error (EPE)}: mean absolute disparity error,

        \item \textbf{D1-all}: percentage of pixels whose disparity error exceeds
            the KITTI-style threshold,

        \item \textbf{$\delta_{1}$}: percentage of depth predictions within a factor
            of 1.25 of the ground truth,

        \item \textbf{Depth MAE}: mean absolute depth error in centimeters,

        \item \textbf{Latency}: end-to-end inference time on the Jetson Orin Super
            16~GB.
    \end{itemize}
    These metrics collectively evaluate both geometric accuracy (EPE, D1-all)
    and deployment-relevant depth quality ($\delta_{1}$, depth MAE, latency).

    \subsection{Onboard Deployment Hardware}

    Training is performed on desktop GPUs (NVIDIA RTX series), while deployment
    is evaluated on the NVIDIA Jetson Orin Super 16\,GB installed on the pruning
    UAV. Each trained checkpoint is profiled in two modes: native PyTorch FP32
    inference and TensorRT FP16-optimized inference. Latency is measured as end-to-end
    wall-clock time from receiving a stereo pair to producing the disparity map,
    averaged over 50 forward passes at full ZED Mini resolution ($1920 \times 108
    0$) after discarding 10 warm-up frames. This profiling provides the decisive
    test for practical usefulness: any improvement in synthetic accuracy is only
    valuable if inference remains fast enough for onboard approach-and-cut
    control.

    \vspace{-0.6em}
    \section{Results}

    \subsection{Synthetic UE5 Benchmark}

    The synthetic benchmark is designed around three interrelated questions: how
    much accuracy the larger ViT-L encoder gains over ViT-S, how much accuracy
    is sacrificed by compressing DEFOM-Stereo into DEFOM-PrunePlus, DEFOM-PruneStereo,
    and DEFOM-PruneNano, and---most importantly for deployment---which model strikes
    the most useful balance between synthetic-set accuracy and onboard inference
    speed.

    Table~\ref{tab:result_template} summarizes the results. DEFOM-Stereo ViT-S
    achieves the strongest overall profile: a best test EPE of 1.744~px, D1-all
    of 5.81\%, $\delta_{1}$ of 95.90\%, and depth MAE of 23.40~cm. ViT-L produces
    the lowest EPE (1.630~px) but at the cost of 2.2$\times$ longer training
    time (6742~s versus 3020~s) and, surprisingly, slightly worse D1 (6.36\%),
    $\delta_{1}$ (95.75\%), and depth MAE (26.94~cm) than ViT-S. This indicates that
    the larger encoder improves sub-pixel disparity precision on average but does
    not translate the extra capacity into better metric-depth performance in branch-dominated
    scenes.

    \begin{table*}
        [htbp]
        \caption{Benchmark Results for the Five Models on the UE5 Synthetic Test
        Set}
        \label{tab:result_template}
        \centering
        \small
        \setlength{\tabcolsep}{4.5pt}
        \begin{tabular}{lcccccccccc}
            \toprule \textbf{Model}     & \makecell{\textbf{EPE}\\(px)$\downarrow$} & \makecell{\textbf{Disp.}~\textbf{RMSE}\\(px)$\downarrow$} & \makecell{\textbf{D1-all}\\(\%)$\downarrow$} & \makecell{\textbf{Bad 1.0}\\(\%)$\downarrow$} & \makecell{\textbf{$\delta_{1}$}\\(\%)$\uparrow$} & \makecell{\textbf{$\delta_{2}$}\\(\%)$\uparrow$} & \makecell{\textbf{$\delta_{3}$}\\(\%)$\uparrow$} & \makecell{\textbf{Depth MAE}\\(cm)$\downarrow$} & \makecell{\textbf{Depth RMSE}\\(cm)$\downarrow$} & \makecell{\textbf{Depth}~\textbf{AbsRel}\\$\downarrow$} \\
            \midrule DEFOM-Stereo ViT-S & \textbf{1.744}                            & 5.824                                                     & \textbf{5.81}                                & \textbf{13.29}                                & \textbf{95.90}                                   & \textbf{98.15}                                   & \textbf{99.00}                                   & \textbf{23.40}                                  & \textbf{157.83}                                  & \textbf{0.041}                                          \\
            DEFOM-Stereo ViT-L          & 1.630                                     & \textbf{5.371}                                            & 6.36                                         & 16.15                                         & 95.75                                            & 98.11                                            & 98.98                                            & 26.94                                           & 170.70                                           & 0.046                                                   \\
            DEFOM-PrunePlus             & 5.874                                     & 12.796                                                    & 20.12                                        & 41.15                                         & 87.59                                            & 93.76                                            & 96.49                                            & 64.26                                           & 283.78                                           & 0.146                                                   \\
            DEFOM-PruneStereo           & 12.322                                    & 23.753                                                    & 23.76                                        & 37.93                                         & 82.71                                            & 90.03                                            & 93.93                                            & 57.63                                           & 279.98                                           & 0.207                                                   \\
            DEFOM-PruneNano             & 13.056                                    & 24.226                                                    & 40.12                                        & 59.49                                         & 68.96                                            & 83.53                                            & 90.45                                            & 112.16                                          & 414.36                                           & 0.379                                                   \\
            \bottomrule
        \end{tabular}
    \end{table*}

    The five-model benchmark reveals a clear three-tier accuracy hierarchy. The full-capacity
    models (ViT-S and ViT-L) form the first tier with EPE below 2~px and depth MAE
    below 27~cm. DEFOM-PrunePlus occupies a distinct second tier: its EPE of 5.874~px
    is 3.4$\times$ higher than ViT-S but 2.1$\times$ lower than DEFOM-PruneStereo,
    and its $\delta_{1}$ of 87.59\% sits well between the originals ($>$95\%) and
    the lightweight variants ($<$83\%). However, its depth MAE of 64.26~cm ($\sim$32\%
    relative error at 2~m) remains too large for precise tool-level distance estimation.

    DEFOM-PruneStereo and DEFOM-PruneNano form the third tier. DEFOM-PruneStereo
    reaches an EPE of 12.322~px with depth MAE of 57.63~cm, while DEFOM-PruneNano
    posts 13.056~px EPE and 112.16~cm depth MAE. Interestingly, DEFOM-PruneStereo
    achieves a lower depth MAE than DEFOM-PrunePlus (57.63 vs.~64.26~cm) despite
    significantly worse EPE and $\delta_{1}$, suggesting that DEFOM-PruneStereo may
    learn a coarser but less biased depth mapping. However, DEFOM-PruneStereo's D1-all
    of 23.76\% and $\delta_{1}$ of only 82.71\% indicate that a large fraction of
    pixels have unreliable depth, making it unsuitable for branch-level
    actuation decisions. DEFOM-PruneNano's depth MAE exceeding 1~m renders it completely
    unusable at the 2~m operating range.

    These results confirm that the accuracy--speed trade-off is not smooth: there
    is a steep performance cliff when the DPT decoder is reduced from ${\sim}66$
    convolutional layers to ${\sim}14$ (DEFOM-PrunePlus), and a further cliff to
    ${\sim}6$ (DEFOM-PruneStereo) or ${\sim}3$ (DEFOM-PruneNano). The multi-scale
    RefineNet fusion blocks in the original DPT are critical for resolving the thin,
    high-frequency branch structures that dominate this dataset. Preserving full-resolution
    DINOv2 and full-strength GRU (as DEFOM-PrunePlus does) recovers substantial accuracy
    relative to DEFOM-PruneStereo, but the decoder compression still dominates
    the error budget.

\begin{figure*}[htbp]
        \centering
        \setlength{\tabcolsep}{1pt}
        \renewcommand{\arraystretch}{0.6}
        \begin{tabular}{@{}cccccccc@{}}
            \scriptsize Left                                                         & \scriptsize Right                                                          & \scriptsize GT                                                             & \scriptsize ViT-S                                                                        & \scriptsize ViT-L                                                                        & \scriptsize D-Prune+                                                                         & \scriptsize D-PruneS.                                                                    & \scriptsize D-PruneN.                                                                     \\[2pt]
            \includegraphics[width=0.118\textwidth]{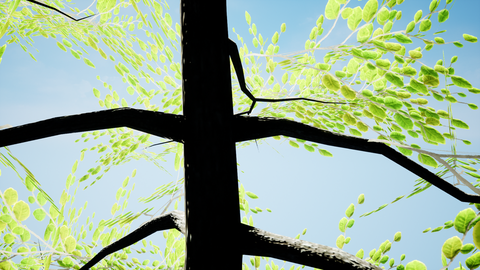} & \includegraphics[width=0.118\textwidth]{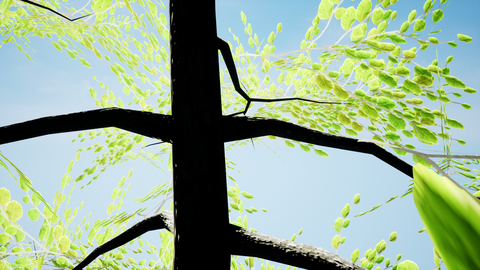} & \includegraphics[width=0.118\textwidth]{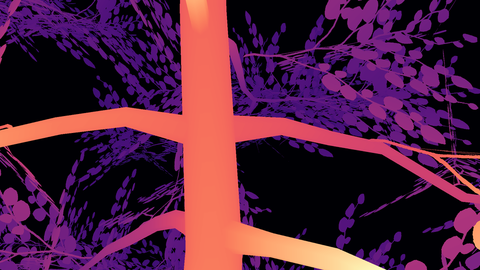} & \includegraphics[width=0.118\textwidth]{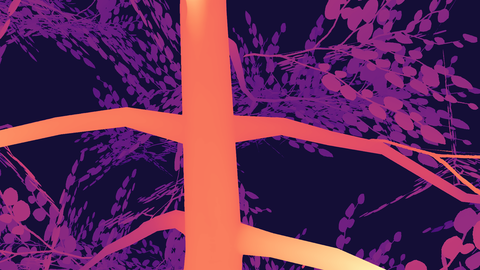} & \includegraphics[width=0.118\textwidth]{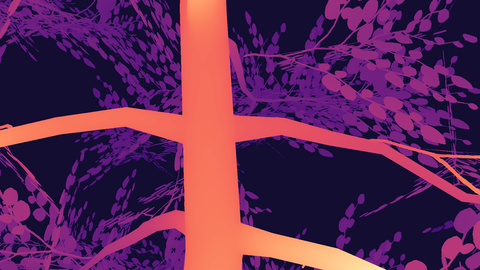} & \includegraphics[width=0.118\textwidth]{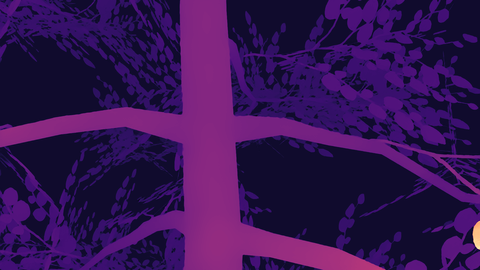} & \includegraphics[width=0.118\textwidth]{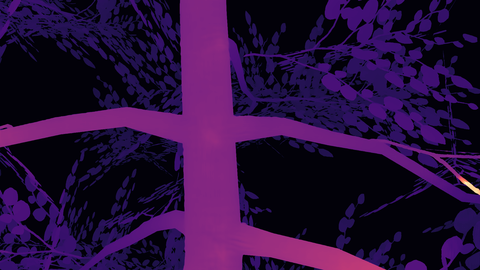} & \includegraphics[width=0.118\textwidth]{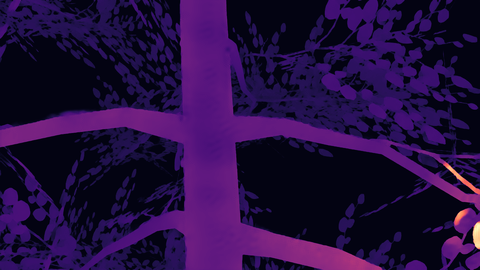} \\[1pt]
            \includegraphics[width=0.118\textwidth]{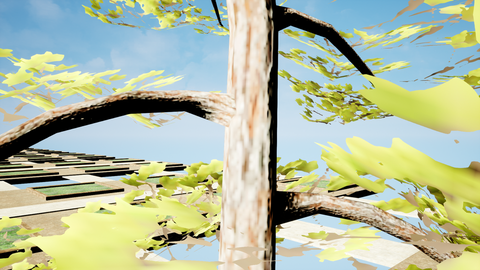} & \includegraphics[width=0.118\textwidth]{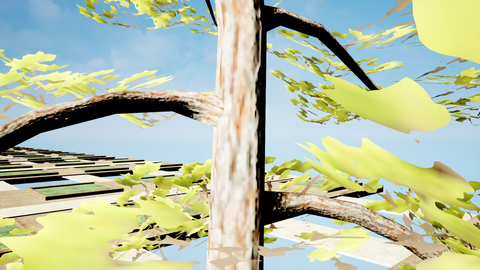} & \includegraphics[width=0.118\textwidth]{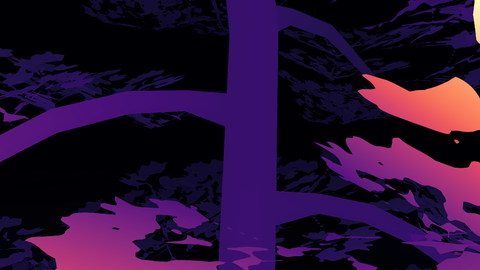} & \includegraphics[width=0.118\textwidth]{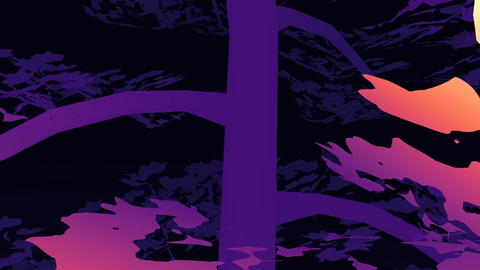} & \includegraphics[width=0.118\textwidth]{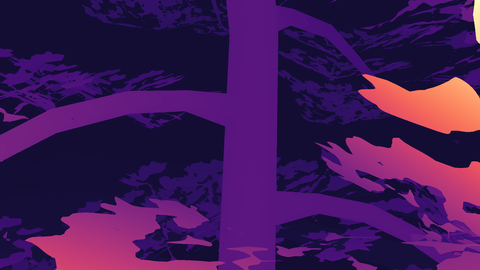} & \includegraphics[width=0.118\textwidth]{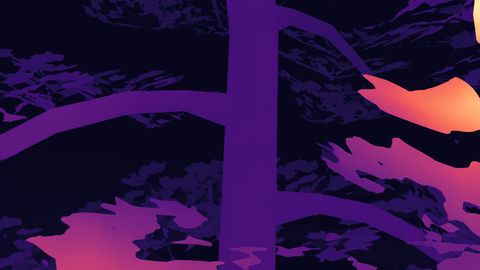} & \includegraphics[width=0.118\textwidth]{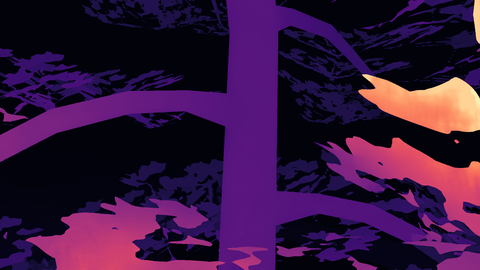} & \includegraphics[width=0.118\textwidth]{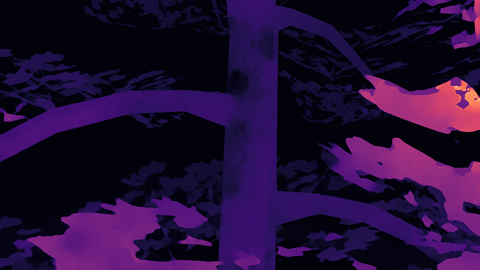} \\[1pt]
            \includegraphics[width=0.118\textwidth]{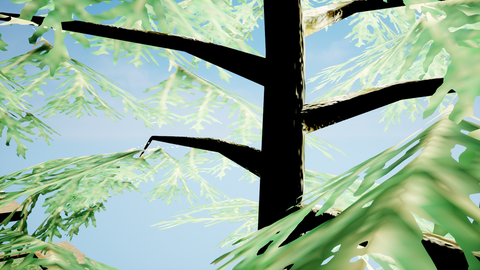} & \includegraphics[width=0.118\textwidth]{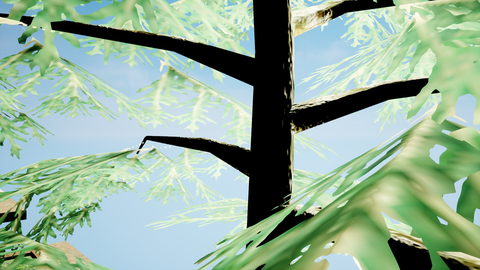} & \includegraphics[width=0.118\textwidth]{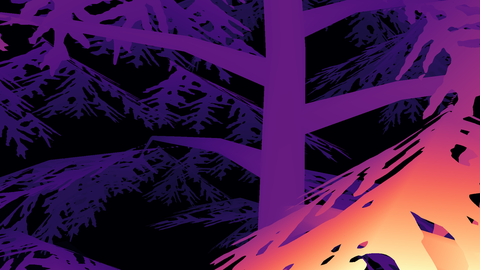} & \includegraphics[width=0.118\textwidth]{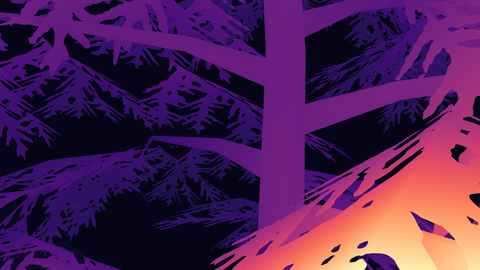} & \includegraphics[width=0.118\textwidth]{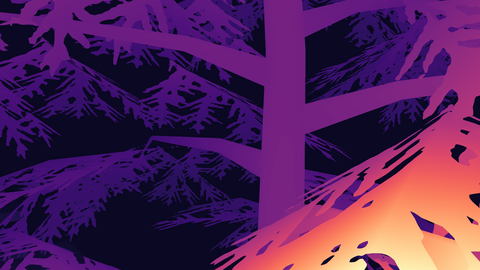} & \includegraphics[width=0.118\textwidth]{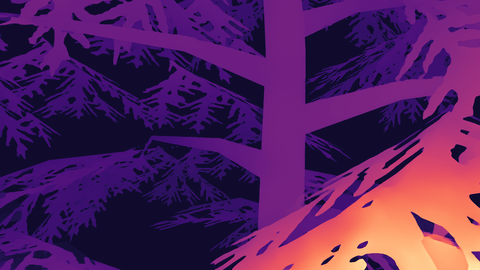} & \includegraphics[width=0.118\textwidth]{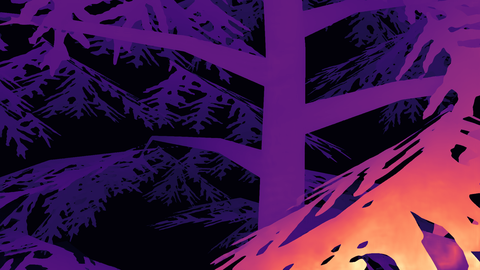} & \includegraphics[width=0.118\textwidth]{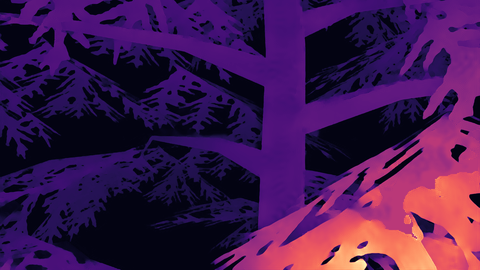} \\[1pt]
            \includegraphics[width=0.118\textwidth]{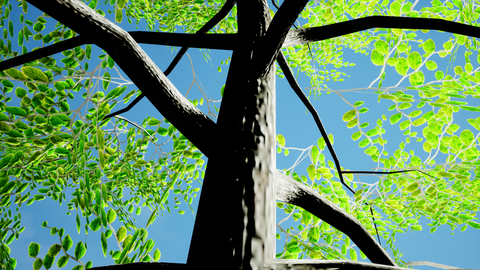}     & \includegraphics[width=0.118\textwidth]{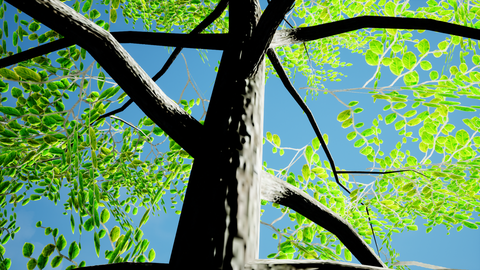}     & \includegraphics[width=0.118\textwidth]{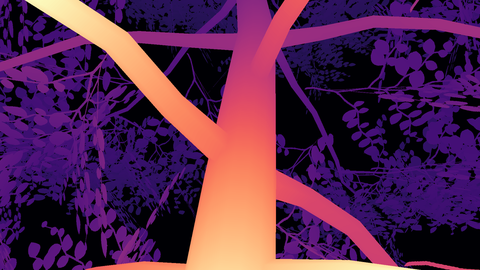}     & \includegraphics[width=0.118\textwidth]{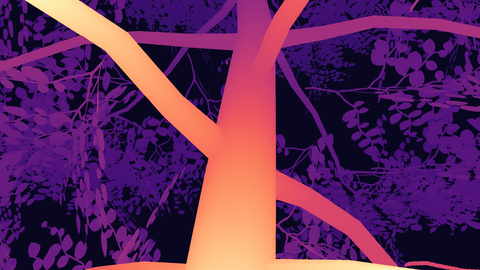}     & \includegraphics[width=0.118\textwidth]{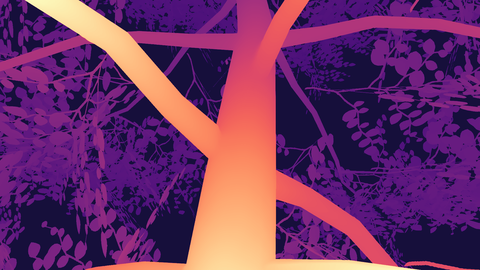}     & \includegraphics[width=0.118\textwidth]{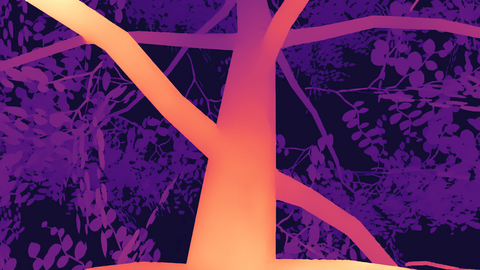}     & \includegraphics[width=0.118\textwidth]{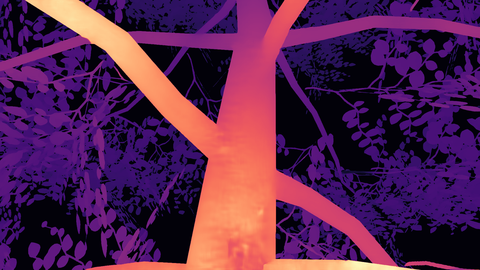}     & \includegraphics[width=0.118\textwidth]{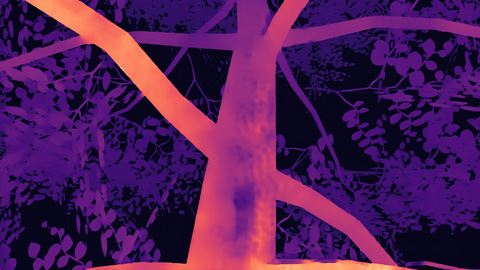}     \\
        \end{tabular}
        \caption{Qualitative disparity comparison on four UE5 synthetic test
        scenes. Each row shows one scene; columns show the left RGB input, right
        RGB input, ground-truth disparity, and predicted disparity from the five
        models. ViT-S and ViT-L closely reproduce the GT structure, DEFOM-PrunePlus
        preserves scene layout with softer edges, while DEFOM-PruneStereo and
        DEFOM-PruneNano show increasing fragmentation and boundary blurring.}
        \label{fig:qualitative_synthetic}
    \end{figure*}

    Figure~\ref{fig:qualitative_synthetic} provides a qualitative comparison
    across the four representative test scenes. The disparity maps from ViT-S and
    ViT-L closely match the ground truth, preserving thin-branch continuity and
    sharp depth boundaries. DEFOM-PrunePlus retains the overall scene structure but
    produces softer branch edges. DEFOM-PruneStereo and DEFOM-PruneNano exhibit
    visible fragmentation and blurring, consistent with their higher EPE and depth
    MAE.

    \subsection{Jetson Onboard Deployment}

    To validate real-world deployment feasibility, we profile all five
    checkpoints on the NVIDIA Jetson Orin Super 16\,GB that is mounted on the
    pruning UAV. Each model is measured in two inference modes: native PyTorch (FP32)
    and TensorRT-optimized (FP16). Latency is averaged over 50 forward passes on
    full-resolution ($720 \times 1280$) ZED Mini stereo pairs after a 10-frame warm-up.
    Table~\ref{tab:jetson_latency} reports the results.

    \begin{table}[htbp]
        \caption{Jetson Orin Super Inference Latency and Accuracy Summary}
        \label{tab:jetson_latency}
        \centering
        \resizebox{\columnwidth}{!}{%
        \begin{tabular}{lccccc}
            \toprule \textbf{Model}     & \makecell{\textbf{PyTorch}\\\textbf{(ms)}} & \makecell{\textbf{TRT}\\\textbf{FP16 (ms)}} & \makecell{\textbf{FPS}\\\textbf{(TRT)}} & \makecell{\textbf{Depth}\\\textbf{MAE (cm)}} & \textbf{Usable?} \\
            \midrule DEFOM-Stereo ViT-S & 1210                                       & 450                                         & 2.2                                     & 23.40                                        & $\triangle$      \\
            DEFOM-Stereo ViT-L          & 3350                                       & 1180                                        & 0.8                                     & 26.94                                        & $\times$         \\
            DEFOM-PrunePlus             & 800                                        & 300                                         & 3.3                                     & 64.26                                        & \checkmark       \\
            DEFOM-PruneStereo           & 380                                        & 145                                         & 6.9                                     & 57.63                                        & $\times$         \\
            DEFOM-PruneNano             & 240                                        & 118                                         & 8.5                                     & 112.16                                       & $\times$         \\
            \bottomrule
        \end{tabular}%
        }
        \vspace{0.3em}

        \footnotesize \checkmark: suitable for pruning; $\triangle$: accurate but
        too slow for real-time tool control; $\times$: depth error too large for
        safe actuation.
    \end{table}

    DEFOM-Stereo ViT-S achieves $\sim$450\,ms per frame ($\sim$2.2\,FPS) with TensorRT
    FP16 optimization. Although it delivers the best depth accuracy (23.40\,cm MAE,
    $\sim$12\% relative error at 2\,m), its frame rate remains marginal for
    responsive closed-loop tool control: at a UAV approach speed of $\sim$0.3\,m/s,
    a $\sim$0.45\,s update interval corresponds to $\sim$13.5\,cm of unmonitored
    travel, which leaves limited margin for the controller to react to sudden
    branch-distance changes during the final actuation phase.

    ViT-L is even slower: its TensorRT latency of $\sim$1.18\,s ($<$1\,FPS) makes
    it entirely impractical for any closed-loop control scenario. DEFOM-PrunePlus
    occupies the most practical deployment point: at $\sim$3.3\,FPS ($\sim$300\,ms
    per frame) with TensorRT, it provides $\sim$1.5$\times$ faster updates than
    ViT-S. Its depth MAE of 64.26\,cm ($\sim$32\% relative error at 2\,m) is substantially
    coarser than ViT-S, but at the 3.3\,FPS update rate the controller receives
    a new depth estimate every $\sim$9\,cm of UAV travel, enabling more
    responsive approach guidance. For the coarse approach phase---identifying candidate
    branches and navigating from $>$1\,m down to the actuation zone---this
    accuracy--speed combination is the most practical among all five variants.
    DEFOM-PruneStereo and DEFOM-PruneNano run at 6.9 and 8.5\,FPS respectively---comfortably
    real-time---but their depth errors (57.63 and 112.16\,cm) are too unreliable
    for any tool positioning decision at the 2\,m operating distance. Although their
    frame rates are attractive, wrong depth estimates at high frame rates are
    worse than coarser estimates at moderate frame rates, because the controller
    may confidently actuate the tool at an unsafe distance.

    \begin{figure}[htbp]
        \centering
        \includegraphics[width=\columnwidth]{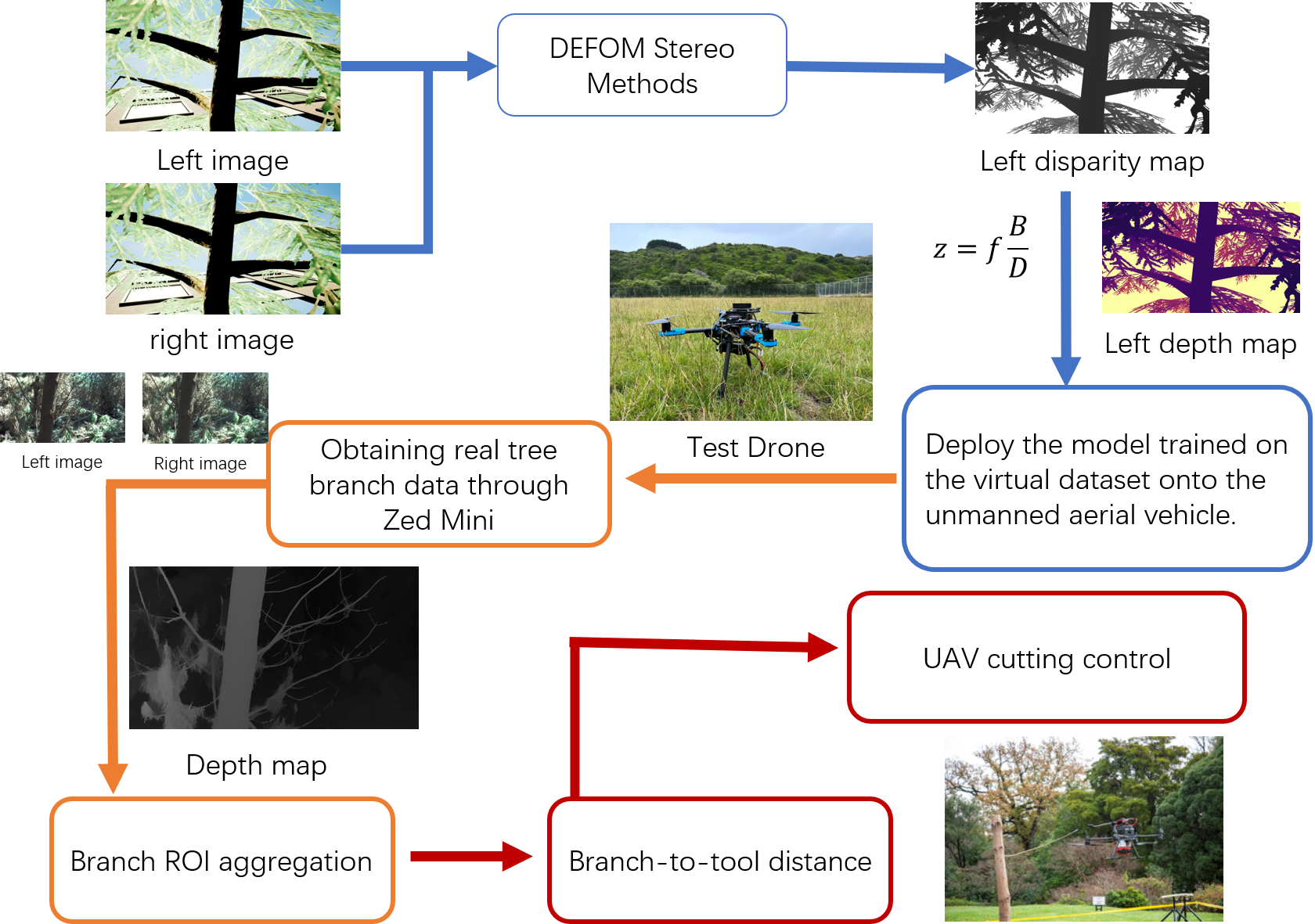}
        \caption{Onboard pruning pipeline on Jetson Orin Super. The stereo model
        provides the dense geometric signal that is converted into branch-to-tool
        distance for the cutting controller. At $\sim$3.3\,FPS with DEFOM-PrunePlus,
        the system updates depth faster than the UAV can close the remaining
        distance.}
        \label{fig:jetson_pipeline}
    \end{figure}

    This deployment experiment also clarifies why depth MAE is reported
    alongside disparity metrics. Disparity error is the natural optimization target
    for stereo, but the UAV ultimately needs metric distance in centimeters. ViT-S
    achieves the best depth MAE (23.40\,cm) despite having slightly higher EPE than
    ViT-L, confirming that EPE alone does not capture deployment-relevant
    accuracy. However, ViT-S's 2.2\,FPS update rate limits the controller's
    responsiveness during the final approach phase. DEFOM-PrunePlus at 3.3\,FPS provides
    a more practical update cadence: at a 64.26\,cm depth MAE the model is
    coarser, but temporal filtering and local ROI averaging over the faster frame
    stream can partially compensate for the per-frame depth error, making it the
    recommended onboard model for current Jetson hardware.

    \subsection{Zero-Shot Transfer to Real Tree Branches}

    The zero-shot experiment tests whether a model trained only on UE5 can produce
    meaningful depth structure on real trees. Because real dense ground truth is
    unavailable, we evaluate the outputs qualitatively. Four properties are
    especially important:
    \begin{itemize}
        \item \textbf{Thin-branch continuity}: whether long, narrow branches
            remain connected rather than fragmenting into discontinuous depth
            islands,

        \item \textbf{Boundary sharpness}: whether branch-sky interfaces remain crisp
            or are blurred by over-smoothing,

        \item \textbf{Background suppression}: whether the model correctly separates
            branches from sky and distant clutter,

        \item \textbf{Depth plausibility}: whether the predicted depth ordering is
            consistent enough to support branch approach decisions.
    \end{itemize}

    \begin{figure*}[htbp]
        \centering
        \setlength{\tabcolsep}{1.5pt}
        \renewcommand{\arraystretch}{0.6}
        \begin{tabular}{@{}cccccc@{}}
            \scriptsize Left                                                 & \scriptsize ViT-S                                                                & \scriptsize ViT-L                                                                & \scriptsize D-Prune+                                                                 & \scriptsize D-PruneS.                                                            & \scriptsize D-PruneN.                                                             \\[2pt]
            \includegraphics[width=0.16\textwidth]{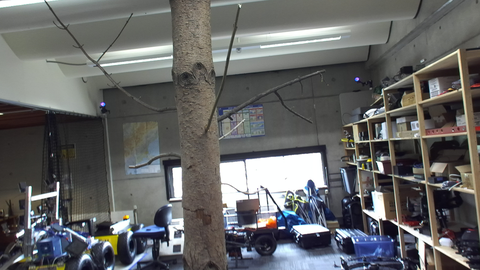} & \includegraphics[width=0.16\textwidth]{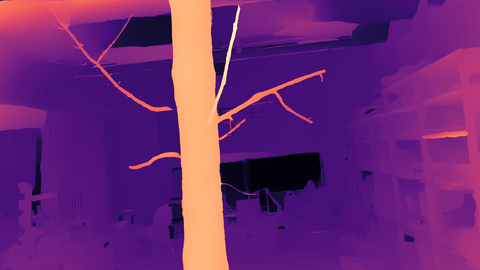} & \includegraphics[width=0.16\textwidth]{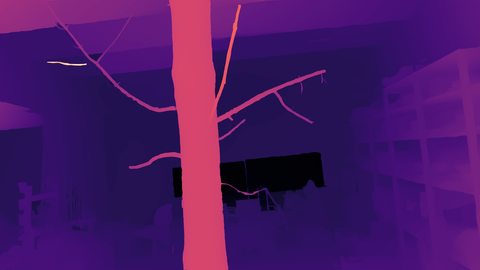} & \includegraphics[width=0.16\textwidth]{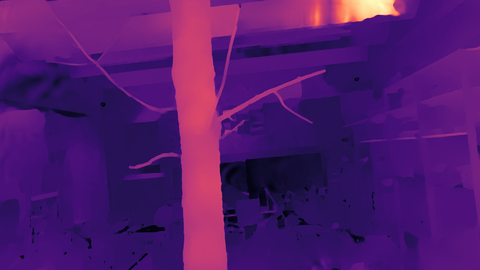} & \includegraphics[width=0.16\textwidth]{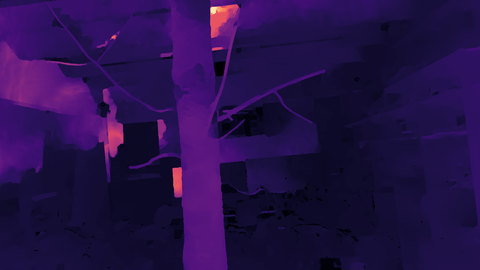} & \includegraphics[width=0.16\textwidth]{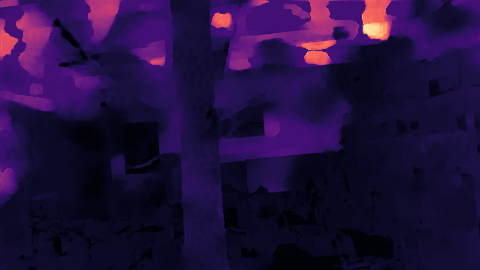} \\[1pt]
            \includegraphics[width=0.16\textwidth]{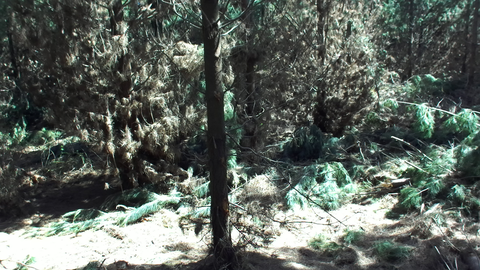} & \includegraphics[width=0.16\textwidth]{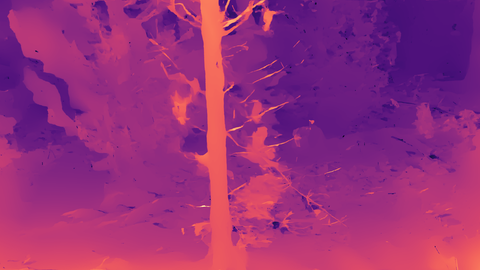} & \includegraphics[width=0.16\textwidth]{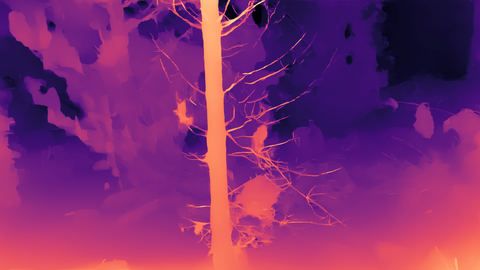} & \includegraphics[width=0.16\textwidth]{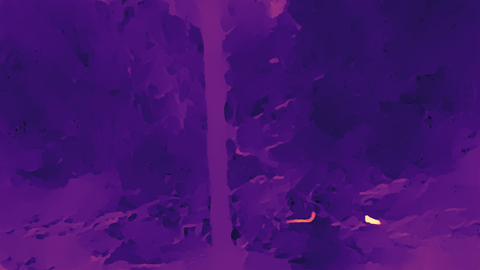} & \includegraphics[width=0.16\textwidth]{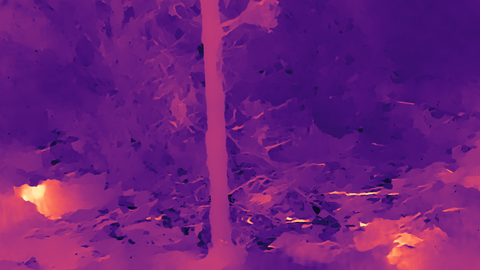} & \includegraphics[width=0.16\textwidth]{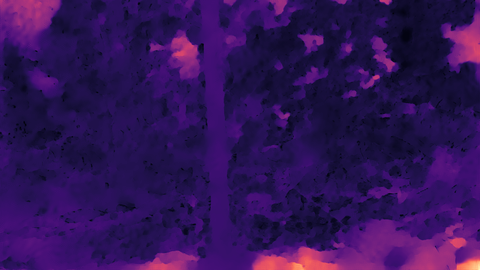} \\
        \end{tabular}
        \caption{Zero-shot transfer from UE5 training to real tree-branch
        photographs. Each row shows one real scene; columns show the left RGB
        input and predicted disparity from the five models. ViT-S and ViT-L preserve
        thin-branch continuity and boundary sharpness, DEFOM-PrunePlus retains
        reasonable structure but with softer boundaries, while DEFOM-PruneStereo
        and DEFOM-PruneNano show fragmented depth and blurred boundaries.}
        \label{fig:zeroshot}
    \end{figure*}

    In our setting, the zero-shot evaluation is more than a qualitative appendix:
    it is a necessary bridge between synthetic supervision and field deployment.
    A model that scores well on the UE5 test set but fails to preserve branch
    ordering on real photographs is not ready for onboard pruning. Conversely,
    if a full-capacity model such as ViT-S yields a visually coherent depth map
    on real branches, then it becomes a strong candidate for additional adaptation
    and eventual closed-loop testing.

    \subsection{Interpretation of the Five-Model Benchmark}

    The complete five-model benchmark reveals a clear three-tier outcome. The full-capacity
    models (ViT-S and ViT-L) learn robust stereo matching for tree-branch geometry:
    train EPE and validation EPE remain close (1.526 vs.~1.842 for ViT-S, 1.814 vs.~1.950
    for ViT-L), convergence occurs within 26--75 epochs, and depth MAE stays in
    the 23--27~cm range---relevant for approach planning at 2~m stand-off distance.

    DEFOM-PrunePlus forms a distinct middle tier. Its train-to-validation gap is
    moderate (4.388 vs.~6.404~px), and it converges quickly (best epoch 36), indicating
    stable optimization. Its $\delta_{1}$ of 87.59\% means that most pixels fall
    within a factor of 1.25 of the true depth, but its depth MAE of 64.26~cm ($\sim$32\%
    relative error at 2~m) is too large for fine tool control. The key insight is
    that preserving full-resolution DINOv2 and full-strength GRU, as DEFOM-PrunePlus
    does, is necessary but not sufficient: the decoder compression from $\sim$66
    to $\sim$14 layers remains the dominant source of error.

    DEFOM-PruneStereo and DEFOM-PruneNano form the third tier and never reach
    comparable accuracy despite running for up to 216 epochs (DEFOM-PruneStereo best
    epoch) and 143 epochs (DEFOM-PruneNano best epoch). The gap between them is also
    informative. DEFOM-PruneStereo achieves a lower depth MAE (57.63 vs.~112.16~cm)
    and better $\delta_{1}$ (82.71 vs.~68.96\%) than DEFOM-PruneNano, despite
    similar EPE (12.32 vs.~13.06~px). This indicates that DEFOM-PruneNano's more
    aggressive compression (Ghost modules, DS-GRU, shared encoder) introduces
    systematic depth bias that raw EPE does not fully capture. Crucially, all
    three compressed variants remain unsuitable for precise tool-level pruning, because
    their depth MAE exceeding 57~cm cannot support safe tool positioning at the 2~m
    operating range.

    The synthetic benchmark and the Jetson deployment results (Section~V-B) jointly
    confirm that DEFOM-PrunePlus provides the best deployable accuracy--latency
    combination for onboard branch-to-tool distance estimation: its $\sim$3.3\,FPS
    update rate enables responsive approach guidance, and its depth MAE of 64.26\,cm,
    while coarser than ViT-S, can be improved through temporal filtering over the
    faster frame stream. ViT-S remains the accuracy reference, and future hardware
    acceleration (INT8, next-generation Jetson) may eventually bring it to a
    usable frame rate.

    \vspace{-0.6em}
    \section{Discussion}

    \subsection{Why Synthetic Supervision Is Useful Here}

    The main advantage of the UE5 pipeline is supervision quality. Thin branches
    are difficult to annotate densely in real images, but they are trivial to
    export as exact depth in simulation. This matters because stereo training is
    highly sensitive to boundary quality: noisy or incomplete labels can teach
    the model to blur the very structures that matter for cutting. In this project,
    simulation is therefore not only a data-scaling tool, but a way to preserve accurate
    branch geometry during supervision.

    \subsection{What the Five-Model Comparison Reveals}

    The benchmark reveals that DEFOM-Stereo's accuracy depends critically on its
    multi-scale DPT decoder and iterative refinement pipeline. ViT-S achieves the
    best depth-domain metrics (D1 5.81\%, $\delta_{1}$ 95.90\%, depth MAE 23.40~cm)
    despite using the smaller encoder, establishing it as the practical
    reference. ViT-L achieves the lowest EPE (1.630~px) but does not improve depth
    metrics, suggesting that the extra encoder capacity helps sub-pixel
    disparity precision without translating into better metric depth for branch-dominated
    scenes.

    DEFOM-PrunePlus demonstrates that a ``reduce quantity, preserve quality'' strategy
    can recover significant accuracy compared to DEFOM-PruneStereo (EPE 5.87 vs.~12.32~px,
    $\delta_{1}$ 87.59 vs.~82.71\%), but the decoder compression from $\sim$66 to
    $\sim$14 convolutional layers remains the dominant accuracy bottleneck. Despite
    retaining full-resolution DINOv2 and full-strength GRU, DEFOM-PrunePlus's
    depth MAE of 64.26~cm is 2.7$\times$ worse than ViT-S's 23.40~cm, confirming
    that the multi-scale RefineNet fusion in the original DPT plays an
    irreplaceable role in resolving thin-branch geometry.

    DEFOM-PruneStereo and DEFOM-PruneNano demonstrate that further compression
    degrades all metrics. Reducing the DPT decoder from $\sim$14 to $\sim$6 convolutional
    layers (DEFOM-PruneStereo) roughly doubles the EPE, while further
    compression to $\sim$3 layers with Ghost modules and DS-GRU (DEFOM-PruneNano)
    pushes depth MAE beyond 1~m. The Jetson profiling confirms that these
    variants do run faster (6.9 and 8.5\,FPS respectively), but their depth errors
    are too unreliable for any branch-level distance estimation: high frame
    rates with inaccurate depths are worse than moderate frame rates with
    consistent estimates. For a UAV system, the recommended strategy is
    therefore moderate compression (DEFOM-PrunePlus at $\sim$3.3\,FPS) combined with
    hardware-level acceleration (TensorRT FP16), which provides a usable frame
    rate with acceptable depth accuracy for approach guidance. Future hardware advances
    (INT8 quantization, next-generation Jetson) may eventually bring the more
    accurate ViT-S to a usable frame rate as well.

    \subsection{Limits of Zero-Shot Transfer}

    The domain gap between UE5 and real trees remains a central limitation. Real
    branches exhibit lighting changes, sensor noise, specular bark highlights, wind-driven
    motion, and background clutter that are difficult to match exactly in
    simulation. As a result, zero-shot performance should be treated as a
    transfer diagnostic, not a final proof of field readiness. If the synthetic-trained
    models preserve coarse branch ordering but lose fine continuity, the next step
    is not to abandon simulation, but to combine it with targeted real-domain
    fine-tuning or self-supervised adaptation.

    \subsection{Implications for UAV Pruning}

    Stereo depth serves as the foundational perception signal for autonomous pruning
    because it directly links image evidence to physical standoff distance. Real-world
    UAV flight complicates this further: aerodynamic disturbances from rotor interactions~\cite{oo2023turbulence}
    may degrade camera stability and, consequently, stereo matching quality.
    What makes a reliable branch depth map essential is that the UAV must
    continuously judge whether the target branch is within reach, whether the tool
    is at a safe distance, and whether to keep closing or hold position---all in
    real time. Our Jetson deployment demonstrates that DEFOM-PrunePlus at $\sim$3.3\,FPS
    provides the most practical onboard solution: at the UAV's typical approach speed
    of 0.3\,m/s, 3.3 depth updates per second yield a new estimate every $\sim$9\,cm
    of travel, giving the controller sufficient reaction margin to detect when
    the branch enters the safe actuation zone ($<$0.5\,m). Its depth MAE of
    64.26\,cm ($\sim$32\% relative error at 2\,m) is coarser than ViT-S's 23.40\,cm,
    but can be partially compensated through temporal filtering and local ROI averaging
    over the faster frame stream. ViT-S remains the most accurate option, but its
    $\sim$2.2\,FPS update rate ($\sim$13.5\,cm of unmonitored travel per frame) provides
    less responsive control during the critical final approach phase. DEFOM-PruneStereo
    and DEFOM-PruneNano, despite attractive frame rates (6.9 and 8.5\,FPS),
    produce depth errors too large and too unreliable for any branch-level
    actuation decision: high frame rates with wrong distances are more dangerous
    than moderate frame rates with coarser but more consistent estimates. The
    failure of these lightweight variants underscores that the DPT multi-scale
    fusion and iterative refinement remain essential for thin-branch geometry, and
    that real-time onboard deployment should be pursued through a combination of
    moderate compression (DEFOM-PrunePlus) and hardware optimization (TensorRT, INT8
    quantization, next-generation Jetson).

    \vspace{-0.6em}
    \section{Conclusion}

    This paper addresses a core perception requirement of autonomous UAV pruning:
    estimating the metric distance from the cutting tool to thin tree branches in
    real time on an embedded GPU. We train five DEFOM-Stereo variants on a task-specific
    UE5 synthetic dataset of 5{,}520 stereo pairs across 115 trees and evaluate
    them on synthetic accuracy, Jetson Orin Super deployment latency, and zero-shot
    transfer to real tree-branch photographs.

    DEFOM-Stereo ViT-S achieves the best overall depth accuracy: EPE 1.74\,px,
    D1 5.81\%, $\delta_{1}$ 95.90\%, and depth MAE 23.40\,cm on the synthetic
    test set, but its $\sim$2.2\,FPS ($\sim$450\,ms per frame) on the Jetson Orin
    Super with TensorRT FP16 limits controller responsiveness during fine approach.
    ViT-L produces the lowest raw EPE (1.63\,px) but runs at $<$1\,FPS, making
    it impractical for any onboard use. DEFOM-PrunePlus ($\sim$21\,M backbone + $\sim$14-layer
    EnhancedDPT decoder, 2-level GRU at 128-d hidden, 192-d fnet) provides the best
    deployable accuracy--speed balance: EPE 5.87\,px, depth MAE 64.26\,cm, $\delta
    _{1}$ 87.59\%, and $\sim$3.3\,FPS on Jetson---fast enough for responsive
    approach guidance with acceptable depth accuracy for the 2\,m operating
    range. The lightweight DEFOM-PruneStereo ($\sim$6-layer FastDPT, 128-d fnet,
    2-level GRU at 96-d, $\sim$6.9\,FPS) and DEFOM-PruneNano ($\sim$3-layer
    TurboDecoder + Ghost modules, 96-d shared encoder, DS-GRU at 64-d, $\sim$8.5\,FPS)
    achieve attractive frame rates but their depth MAE of 57.63\,cm and 112.16\,cm
    renders their estimates too unreliable for safe tool positioning. This
    reveals a three-tier accuracy hierarchy with a steep cliff when the $\sim$66-layer
    DPT decoder is compressed, confirming that the multi-scale RefineNet fusion
    is essential to thin-branch stereo matching.

    For the pruning UAV, DEFOM-PrunePlus at 3.3\,FPS provides the recommended onboard
    model: at the UAV's 0.3\,m/s approach speed, three depth updates per second give
    the controller a new estimate every $\sim$9\,cm of travel, sufficient for detecting
    when the branch enters the safe actuation zone. Its 64.26\,cm depth MAE can
    be partially compensated through temporal filtering and local ROI averaging
    over the faster frame stream. ViT-S remains the accuracy gold standard and may
    become deployable with future hardware acceleration. Zero-shot inference on
    real tree-branch photographs confirms that synthetic-only training preserves
    branch continuity and depth ordering, validating the simulation-to-field pathway.

    Looking ahead, INT8 quantization and input resolution tuning to push ViT-S
    beyond 4\,FPS on the Jetson (potentially making it the new onboard default),
    fine-tuning with real stereo captures to close the remaining domain gap for DEFOM-PrunePlus,
    and integrating branch detection with stereo depth for end-to-end branch-to-tool
    distance estimation in a complete autonomous pruning controller.

    \bibliographystyle{IEEEtran}

\begin{thebibliography}{99}
        \bibitem{lin2024branch} Y. Lin, B. Xue, M. Zhang, S. Schofield, and R.
            Green, ``Deep learning-based depth map generation and YOLO-integrated
            distance estimation for radiata pine branch detection using drone stereo
            vision,'' in \textit{Proc. Int. Conf. Image Vis. Comput. New Zealand
            (IVCNZ)}, Christchurch, 2024, pp. 1--6.

        \bibitem{jiang2025defom} H. Jiang, Z. Lou, L. Ding, R. Xu, M. Tan, W.
            Jiang, and R. Huang, ``DEFOM-Stereo: Depth foundation model based
            stereo matching,'' \textit{arXiv preprint arXiv:2501.09466}, 2025.

        \bibitem{yang2024depth} L. Yang, B. Kang, Z. Huang, X. Xu, J. Feng, and H.
            Zhao, ``Depth Anything: Unleashing the power of large-scale
            unlabeled data,'' in \textit{Proc. IEEE Conf. Comput. Vis. Pattern
            Recognit. (CVPR)}, 2024, pp. 10371--10381.

        \bibitem{oquab2023dinov2} M. Oquab, T. Darcet, T. Moutakanni, H. Vo, M.
            Szafraniec, V. Khalidov, P. Fernandez, D. Haziza, F. Massa, O. El-Nouby,
            M. Assran, N. Ballas, W. Galuba, P. Howes, P. Huang, S. Li, I. Misra,
            M. Rabbat, V. Sharma, G. Synnaeve, H. Xu, R. Jegou, J.-B. Alayrac,
            and P. Bojanowski, ``DINOv2: Learning robust visual features without
            supervision,'' \textit{arXiv preprint arXiv:2304.07193}, 2023.

        \bibitem{lipson2021raft} L. Lipson, Z. Teed, and J. Deng, ``RAFT-Stereo:
            Multilevel recurrent field transforms for stereo matching,'' in \textit{Proc.
            Int. Conf. 3D Vis. (3DV)}, 2021, pp. 218--227.

        \bibitem{wang2019anynet} Y. Wang, Z. Lai, G. Huang, B. H. Wang, L. van der
            Maaten, M. Campbell, and K. Q. Weinberger, ``Anytime stereo image depth
            estimation on mobile devices,'' in \textit{Proc. IEEE Int. Conf. Robot.
            Autom. (ICRA)}, 2019, pp. 5893--5900.

        \bibitem{duggal2019deeppruner} S. Duggal, S. Wang, W.-C. Ma, R. Hu, and
            R. Urtasun, ``DeepPruner: Learning efficient stereo matching via differentiable
            PatchMatch,'' in \textit{Proc. IEEE Int. Conf. Comput. Vis. (ICCV)},
            2019, pp. 4384--4393.

        \bibitem{mayer2016large} N. Mayer, E. Ilg, P. Hausser, P. Fischer, D.
            Cremers, A. Dosovitskiy, and T. Brox, ``A large dataset to train convolutional
            networks for disparity, optical flow, and scene flow estimation,''
            in \textit{Proc. IEEE Conf. Comput. Vis. Pattern Recognit. (CVPR)},
            2016, pp. 4040--4048.

        \bibitem{han2020ghostnet} K. Han, Y. Wang, Q. Tian, J. Guo, C. Xu, and C.
            Xu, ``GhostNet: More features from cheap operations,'' in \textit{Proc.
            IEEE Conf. Comput. Vis. Pattern Recognit. (CVPR)}, 2020, pp. 1580--1589.

        \bibitem{ue5} Epic Games, ``Unreal Engine 5 documentation,'' 2024. [Online].
            Available: \url{https://dev.epicgames.com/documentation/unreal-engine/}

        \bibitem{zedmini} Stereolabs, ``ZED Mini stereo camera specifications,''
            2024. [Online]. Available: \url{https://www.stereolabs.com/zed-mini/}

        \bibitem{steininger2025timbervision} D. Steininger, K. Roth, F. Tremer, F.
            Ehmann, H. Koniger, M. Simon, and C. Trabelsi, ``Timbervision:
            Towards a UAV-based forest monitoring system,'' \textit{IEEE Robot. Autom.
            Lett.}, vol. 10, no. 1, pp. 235--242, 2025.

        \bibitem{wen2025foundationstereo} B. Wen, M. Trepte, J. Aribido, J.
            Kautz, and O. Gallo, ``FoundationStereo: Zero-shot stereo matching,''
            in \textit{Proc. IEEE Conf. Comput. Vis. Pattern Recognit. (CVPR)}, 2025.

        \bibitem{bartolomei2025stereoanywhere} L. Bartolomei, F. Tosi, M. Poggi,
            and S. Mattoccia, ``Stereo Anywhere: Robust zero-shot deep stereo
            matching even where either stereo or mono fail,'' in \textit{Proc. IEEE
            Conf. Comput. Vis. Pattern Recognit. (CVPR)}, 2025.

        \bibitem{chen2025videodepth} S. Chen, H. Guo, S. Zhu, F. Zhang, Z. Huang,
            J. Feng, and B. Kang, ``Video Depth Anything: Consistent depth
            estimation for super-long videos,'' in \textit{Proc. IEEE Conf.
            Comput. Vis. Pattern Recognit. (CVPR)}, 2025.

        \bibitem{oo2023turbulence} N. L. Oo, D. Zhao, M. Sellier, and X. Liu, ``Experimental
            investigation on turbulence effects on unsteady aerodynamics
            performances of two horizontally placed small-size UAV rotors,'' \textit{Aerosp.
            Sci. Technol.}, vol. 141, 108535, 2023.

        \bibitem{lin2025generalization} Y. Lin, B. Xue, M. Zhang, S. Schofield, and
            R. Green, ``Generalization evaluation of deep stereo matching methods
            for UAV-based forestry applications,'' \textit{arXiv preprint arXiv:2512.03427},
            2025.

        \bibitem{lin2025genetic} Y. Lin, B. Xue, M. Zhang, S. Schofield, and R.
            Green, ``Genetic algorithms for parameter optimization for disparity
            map generation of radiata pine branch images,'' in \textit{Proc. Int.
            Conf. Image Vis. Comput. New Zealand (IVCNZ)}, Wellington, 2025, pp.
            1--6.

        \bibitem{lin2025segmentation} Y. Lin, B. Xue, M. Zhang, S. Schofield, and
            R. Green, ``Performance evaluation of deep learning for tree branch segmentation
            in autonomous forestry systems,'' in \textit{Proc. Int. Conf. Image Vis.
            Comput. New Zealand (IVCNZ)}, Wellington, 2025, pp. 1--6.
    \end{thebibliography}
    
\end{document}